\pdfoutput=1

\documentclass[11pt]{article}

\usepackage[]{ACL2023}

\usepackage{times}
\usepackage{latexsym}

\usepackage[T1]{fontenc}

\usepackage[utf8]{inputenc}

\usepackage{microtype}

\usepackage{inconsolata}

\usepackage{graphicx}
\usepackage{subfigure}

\usepackage{amsmath}
\usepackage{amstext}
\usepackage{amsfonts}
\usepackage{bm}

\usepackage{bbm}
\usepackage{multirow}
\usepackage{booktabs}
\usepackage{array}

\usepackage{soul}

\usepackage{algorithm}
\usepackage{algorithmic}

\usepackage{enumitem}
\setenumerate[1]{itemsep=0pt,partopsep=0pt,parsep=\parskip,topsep=5pt}
\setitemize[1]{itemsep=0pt,partopsep=0pt,parsep=\parskip,topsep=5pt}
\setdescription{itemsep=0pt,partopsep=0pt,parsep=\parskip,topsep=5pt}

\title{Enhancing Continual Relation Extraction via Classifier Decomposition}

\author{{Heming Xia}\textsuperscript{\rm $\dag$,\rm $\ddag$} \quad {Peiyi Wang}\textsuperscript{\rm $\dag$} \quad {Tianyu Liu}\textsuperscript{\rm $\S$} \quad {Binghuai Lin}\textsuperscript{\rm $\S$} \quad {Yunbo Cao}\textsuperscript{\rm $\S$} \quad {Zhifang Sui}\textsuperscript{\rm $\dag$}\\
  \textsuperscript{\rm $\dag$} The MOE Key Laboratory of Computational Linguistics, Peking University\\
  \textsuperscript{\rm $\ddag$} School of Software \& Microelectronics, Peking University \quad
  \textsuperscript{\rm $\S$} Tencent Cloud AI\\
  {\tt \{xiaheming,szf\}@pku.edu.cn \{wangpeiyi9979\}@gmail.com} \\
  {\tt \{rogertyliu, binghuailin, yunbocao\}@tencent.com}}

\begin{document}
\maketitle
\begin{abstract}
Continual relation extraction (CRE) models aim at handling emerging new relations while avoiding catastrophically forgetting old ones in the streaming data. Though improvements have been shown by previous CRE studies, most of them only adopt a vanilla strategy when models first learn representations of new relations. In this work, we point out that there exist two typical biases after training of this vanilla strategy: \textit{classifier bias} and \textit{representation bias}, which causes the previous knowledge that the model learned to be shaded. To alleviate those biases, we propose a simple yet effective classifier decomposition framework that splits the last FFN layer into separated previous and current classifiers, so as to maintain previous knowledge and encourage the model to learn more robust representations at this training stage. Experimental results on two standard benchmarks show that our proposed framework consistently outperforms the state-of-the-art CRE models, which indicates that the importance of the first training stage to CRE models may be underestimated. Our code is available at \url{https://github.com/hemingkx/CDec}.
\end{abstract}

\section{Introduction}
\label{sec:intro}
Continual relation extraction (CRE) \cite{Wang:2019} requires models to learn new relations from a class-incremental data stream while avoiding \textit{catastrophic forgetting} of old relations. To address the problem of catastrophic forgetting, rehearsal-based CRE methods store a few typical instances for each relation on memory and replay the memory data in the subsequent learning process. Despite the simplicity, rehearsal-based methods have become the state-of-the-art CRE methods \cite{Han:2020, Cui:2020, Hu:2022, Wang:2022b}.

Recent rehearsal-based CRE models usually follow a two-stage training paradigm to tackle the \textit{data imbalance} problem\footnote{If models are directly trained with the mixed data which contains sufficient current data and few previous data, severe data imbalance will lead models to overfit on the previous data, which harms the model's performance \citep{Wang:2022a}.}: (1) in \textit{Stage 1}, models are trained with only new data to rapidly learn to identify new relations; (2) in \textit{Stage 2}, models are trained on the updating memory data to alleviate catastrophic forgetting.

\begin{figure}[t]
\centering
\includegraphics[width=1.0\columnwidth]{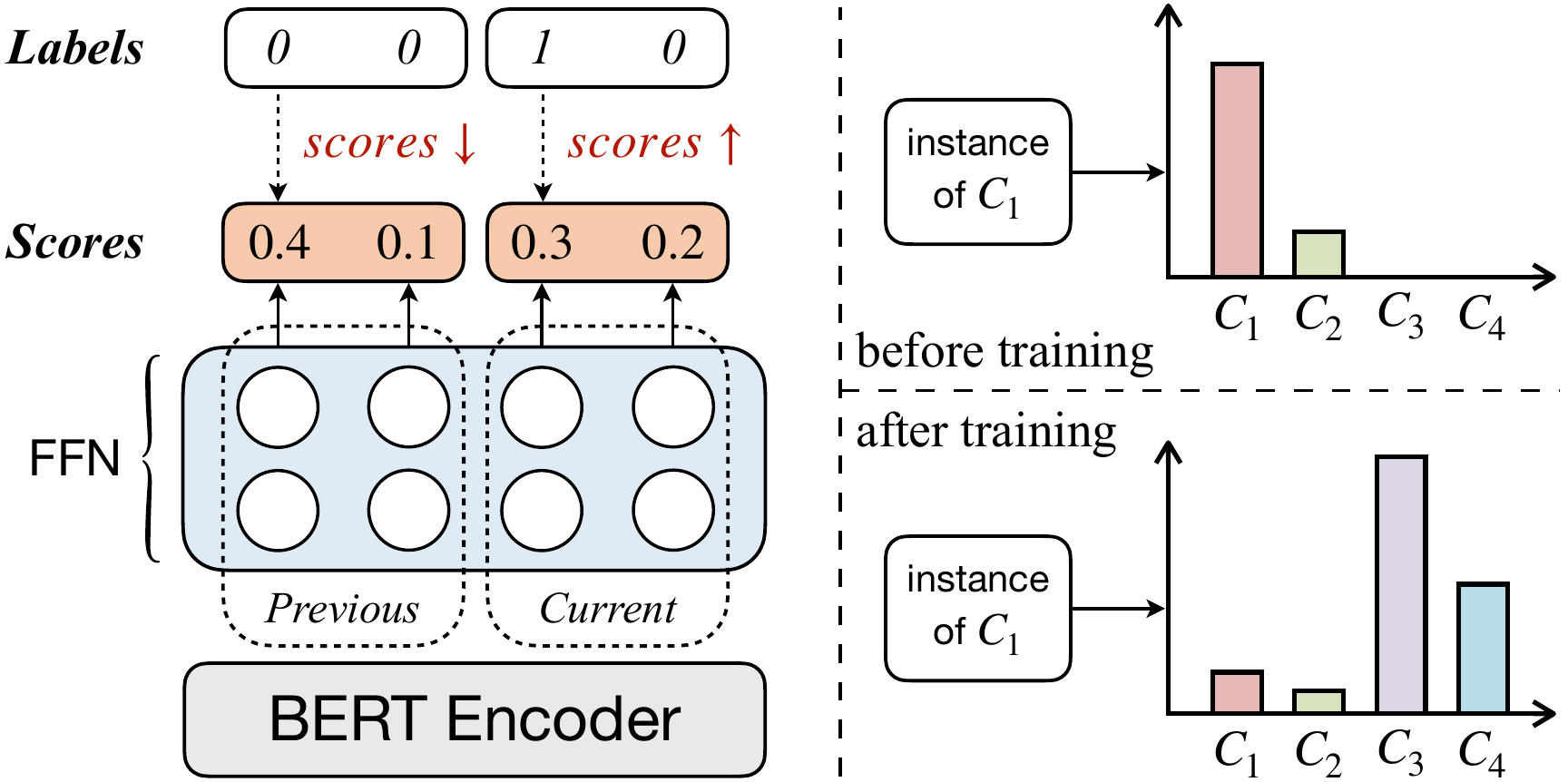}
\caption{A demonstration of the classifier bias in stage 1. Since models are trained with only current data, prediction scores of previous relations are forced to be relatively low after training, i.e., models tend to classify instances into only new relations.} \vspace{-0.3cm}
\label{fig:intro}
\end{figure}

\begin{figure*}[t]
\centering
\vspace{-1.0cm}
\includegraphics[width=1.0\textwidth]{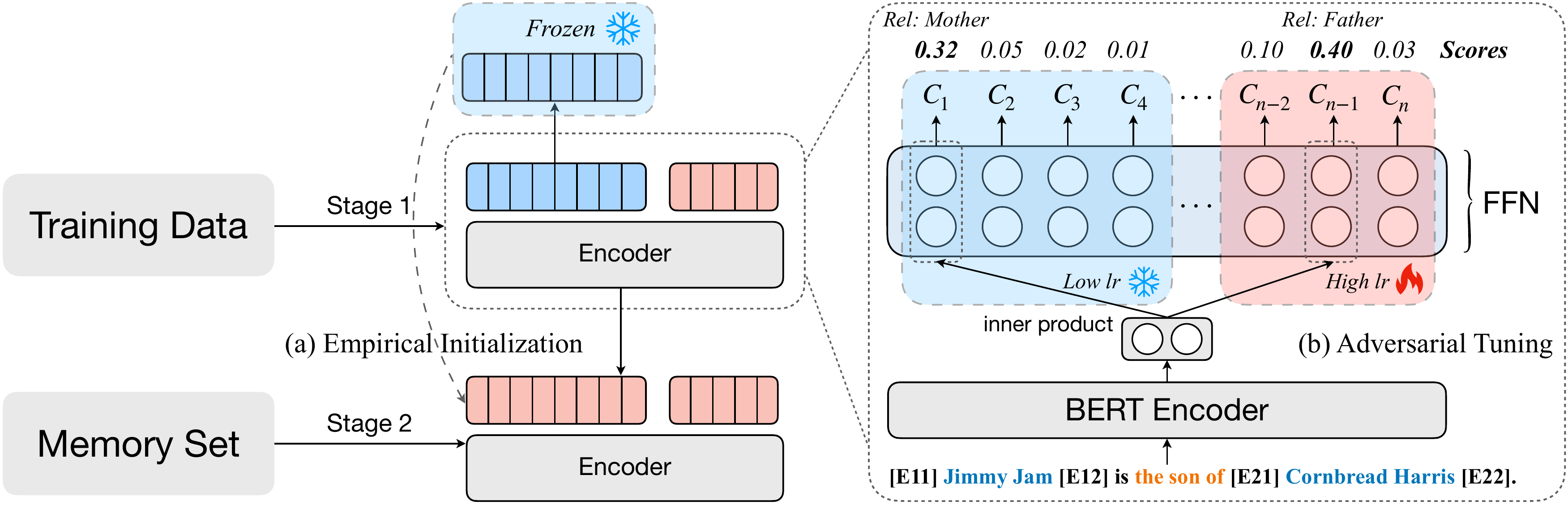}
\caption{An overall demonstration of our proposed classifier decomposition framework, which includes \textbf{(a) Empirical Initialization}: the previous knowledge is frozen and transferred to stage 2 via parameter initialization of the previous classifier; \textbf{(b) Adversarial Tuning}: during stage 1, feedforward nodes in the previous classifier generate adversarial signals, which helps the model learn robust representations.}
\label{fig:Framework}
\end{figure*}

Previous CRE works mainly focus on stage 2, and propose a variety of replay strategies to better utilize the memory data, such as relation prototypes \cite{Han:2020}, memory network \cite{Cui:2020} and contrastive learning \cite{Hu:2022}. However, the exploration of the first training stage is still uncharted territory. In this work, we focus on stage 1 and point out that it suffers from two typical biases that harm the model's performance:
 
\begin{itemize}
    \item \textbf{classifier bias}: without the previous training data, the class weights of previous relations would be improperly updated, leading to a skewed output distribution as shown in Figure \ref{fig:intro}.
    \item \textbf{representation bias}: the learned representations of current relations would be easily overlapped with those of previous relations in the representation space.

\end{itemize}


To alleviate the biases mentioned above, we propose a simple yet effective classifier decomposition framework inspired by \citet{wang2021behind}.
The framework splits the last FFN layer into the separated previous and current classifiers, and
introduces two enhanced strategies: \textit{empirical initialization} and \textit{adversarial tuning}.
Specifically, 
To alleviate the classifier bias,
empirical initialization firstly preserves the well-learned previous classifier nodes and then reuses them after learning current relations.
To ease the representation bias,
adversarial tuning slows down the update of previous classifier weights.
The slowly updating weights can be seen as a series of adversarial signals that induce CRE models to learn distinguishable representations.

To sum up: (1) we find that CRE models suffer from classifier and representation biases when learning the new relations.
(2) we propose a simple yet effective classifier decomposition framework with empirical initialization and adversarial tuning to alleviate these two biases.
(3) Extensive experiments on FewRel and TACRED verify the effectiveness of our method.

\section{Task Formalization}
In CRE, models are trained on a sequence of tasks \{$T_1$, $T_2$, \ldots, $T_k$\}, where the $k$-th task has
its own training set $D_k=\{(x_i, y_i)\}_{i=1}^N$ with $C^k$ new relations. Each task $T_k$ is an independent supervised classification task to identify the instance $x_i$ (including the sentence and entity pair) into its corresponding relation label $y_i$. The goal of CRE is to learn new tasks while avoiding catastrophic forgetting of previously seen tasks. In other words, the model is evaluated to identify an instance into all $C^k_{prev} + C^k$ relations, where $C^k_{prev} = \sum_{i=1}^{k-1}C^i$ is the number of previously seen relations. To alleviate catastrophic forgetting in CRE, previous rehearsal-based work \cite{Han:2020, Cui:2020, Hu:2022, Wang:2022a} adopts an episodic memory module to store a few representative instances for each previous relation. In the subsequent training process, instances in the memory set will be replayed to balance the model's performance on all seen relations.

\begin{table*}[t]
\centering
\small
\vspace{-1.0cm}
\begin{tabular}{l|cccccccccc}
\toprule
\multicolumn{11}{c}{\textbf{FewRel}} \\
\midrule
\textbf{Models} & \textbf{T1} & \textbf{T2} & \textbf{T3} & \textbf{T4} & \textbf{T5} & \textbf{T6} & \textbf{T7} & \textbf{T8} & \textbf{T9} & \textbf{T10}  \\
\midrule
EA-EMR \cite{Wang:2019} & 89.0 & 69.0 & 59.1 & 54.2 & 47.8 & 46.1 &  43.1 & 40.7 & 38.6 & 35.2 \\
CML \cite{Wu:2021} & 91.2 & 74.8 & 68.2 & 58.2 & 53.7 & 50.4 & 47.8 & 44.4 & 43.1 & 39.7 \\
RPCRE \cite{Cui:2020} & 97.9 & 92.7 & 91.6 & 89.2 & 88.4 & 86.8 & 85.1 & 84.1 & 82.2 & 81.5  \\
EMAR$^{\dagger}$ \cite{Han:2020} & 98.2 & 94.1  & 92.0 & 90.8 & 89.7 & 88.1 & 87.2 & 86.1 & 84.8 & 83.6 \\
CRECL$^{\dagger}$ \cite{Hu:2022} & 98.0 & 94.7 & 92.4 & 90.7 & 89.4 & 87.1 & 85.9 & 85.0 & 84.0 & 82.1 \\
CRL \cite{Zhao:2022} & 98.1 & 94.6 & 92.5 & 90.5 & 89.4 & 87.9 & 86.9 & 85.6 & 84.5 & 83.1 \\
\midrule
\textbf{Ours} & \underline{98.2} & \underline{94.9} & \underline{93.2} & \underline{91.9} & \bf{91.3} & \underline{89.6} & \underline{88.3} & \underline{87.1} & \underline{86.0} & \underline{84.6} \\
\quad w/ ACA \cite{Wang:2022b} & \bf{98.4} & \bf{95.4} & \bf{93.2} & \bf{92.1} & \underline{91.0} & \bf{89.7} & \bf{88.3} & \bf{87.4} & \bf{86.4} & \bf{84.8} \\
\toprule
\multicolumn{11}{c}{\textbf{TACRED}} \\
\midrule
\textbf{Models} & \textbf{T1} & \textbf{T2} & \textbf{T3} & \textbf{T4} & \textbf{T5} & \textbf{T6} & \textbf{T7} & \textbf{T8} & \textbf{T9} & \textbf{T10}  \\
\midrule
EA-EMR \cite{Wang:2019} & 47.5 & 40.1 & 38.3 & 29.9 & 24.0 & 27.3 &  26.9 & 25.8 & 22.9 & 19.8\\
CML \cite{Wu:2021} & 57.2 & 51.4 & 41.3 & 39.3 & 35.9 & 28.9 & 27.3 & 26.9 & 24.8 & 23.4 \\
RPCRE \cite{Cui:2020} & 97.6 & 90.6 & 86.1 & 82.4 & 79.8 & 77.2 & 75.1 & 73.7 & 72.4 & 72.4 \\
EMAR$^{\dagger}$ \cite{Han:2020} &\underline{97.8} & 92.4 & 89.6 & 84.6 & 83.2 & 81.3 & 78.7 & 77.1 & 77.3 & 76.8 \\
CRECL$^{\dagger}$ \cite{Hu:2022} & 97.3 & \bf{93.6} & \underline{90.5} & \underline{86.1} & 84.6 & 82.1 & 79.4 & 77.6 & 77.9 & 77.4 \\
CRL \cite{Zhao:2022} & 97.7 & \underline{93.2} & 89.8 & 84.7 & 84.1 & 81.3 & 80.2 & 79.1 & \underline{79.0} & 78.0 \\
\midrule
\textbf{Ours} & \bf{97.9} & 93.1 & 90.1 & 85.8 & \underline{84.7} & \underline{82.6} & \bf{81.0} & \underline{79.6} & \bf{79.5} & \underline{78.6} \\
\quad w/ ACA \cite{Wang:2022b} & 97.7 & 92.8 & \bf{91.0} & \bf{86.7} & \bf{85.2} & \bf{82.9} & \underline{80.8} & \bf{80.2} & 78.8 & \bf{78.6} \\
\bottomrule
\end{tabular}
\caption{Accuracy (\%) on all seen relations at the stage of learning current tasks. $^{\dagger}$ denotes our reproduced results from open source code. Other results are directly taken from \citet{Zhao:2022}. We show the best results in \textbf{boldface} and the second best ones in \underline{underlines}.
}
\label{tab:main}
\end{table*}

\section{Methodology}
The overall of our proposed classifier decomposition framework is illustrated in Figure \ref{fig:Framework}. Following previous work \cite{Hu:2022, Wang:2022a, Zhao:2022}, our model architecture contains two main components: (1) an encoder that generates representations of a given instance and (2) a feed-forward network (FFN) that maps the encoded representation into a probability distribution over all seen relations.

\subsection{Classifier Decomposition}
\label{sec:framework}
To alleviate the illustrated two typical biases (i.e., \textit{classifier bias} and \textit{representation bias}) in the first training stage, we propose a classifier decomposition framework that splits the FFN layer into two independent groups of FFN nodes: the previous nodes and the current ones. The framework contains two enhanced strategies to alleviate the biases in stage 1, which include:

\paragraph{Empirical Initialization} Before the stage 1 training, we keep a frozen copy of the model's previous FFN nodes, which represents the knowledge that the model learned in previous incremental tasks. This frozen copy is then utilized to initialize the parameters of the previous classifier before stage 2 so as to alleviate the classifier bias and retain the previous knowledge.

\paragraph{Adversarial Tuning} To alleviate the representation bias, we propose that the output distribution of previous relations is opposite to the optimization objective of the current task. This distribution can be viewed as adversarial signals, which helps the model learn more unique and distinguishable representations. Taking the analogous relation pair \{$C_1$, $C_{n-1}$\} (e.g., the relation pair of \{``Mother'', ``Father''\}) in Figure \ref{fig:Framework} as an example, given a training instance of $C_{n-1}$, due to the high similarity of encoder representations between $C_1$ and $C_{n-1}$, the prediction score of $C_1$ will also be high. Thus, the training objective of current tasks will lead to two optimization paths: reduce the scores of $C_1$ or force the model to learn more unique representations of $C_{n-1}$. Inspired by this, we propose to slow down the update of previous classifier weights during training so as to steer the learning more toward the second optimization path. Formally,

\begin{equation}
    \theta_{prev} \leftarrow \theta_{prev} - \alpha_{prev} \frac{\partial}{\partial \theta_{prev}} \mathcal{L}
\end{equation}
\begin{equation}
    \theta_{cur} \leftarrow \theta_{cur} - \alpha_{cur} \frac{\partial}{\partial \theta_{cur}} \mathcal{L}
\end{equation}
where $\alpha_{prev}$, $\alpha_{cur}$ are the learning rate of the previous and current output embeddings, respectively. We adopt a lower $\alpha_{prev}$ in the training of stage 1.


\subsection{Training and Inference}
Following previous work \cite{Han:2020, Cui:2020, Hu:2022, Wang:2022a}, the training loss function at task $T_k$ of our framework is given by:
\begin{equation}
    \mathcal{L}=\sum_{i=1}^{D^*}-\log P(y_i|x_i)
\end{equation}
where $P(y_i|x_i)$ is the prediction scores calculated by the FFN layer, $(x_i, y_i)$ is the sample from $D^*$, $D^*$ denotes $D_k$ in stage 1 or the memory set in stage 2. Specifically, $P(y_i|x_i)$ is given by:
\begin{equation}
    P(y_i|x_i) =  {\rm softmax}(\mathbf{W}\mathbf{h})
\end{equation}
where $\mathbf{h} \in \mathbb{R}^d$ is the encoder representation of the instance $x_i$; $\mathbf{W} \in \mathbb{R}^{d \times (C_{prev}^k + C^k})$ stands for the weights of the FFN layer. During inference, we select the relation with the max prediction score as the predicted relation.

\section{Experiments}

\subsection{Experimental Setups}

\paragraph{Datasets} Following previous work \cite{Han:2020, Cui:2020, Hu:2022, Wang:2022a}, our experiments are conducted upon two standard benchmarks, \textbf{FewRel} \cite{Han:2018} and \textbf{TACRED} \cite{Zhang:2017}, please refer to Appendix \ref{sec:datasets} for more details.

\paragraph{Implement Details} For fair comparisons, we use the same experimental settings as \citet{Cui:2020, Zhao:2022}, which randomly divide all relations into 10 sets to simulate 10 tasks and report the average accuracy of 5 different sampling task sequences. The number of stored instances in the memory for each relation is $10$ for all methods.
We adopt the same random seeds to guarantee that the task sequences are exactly the same. We search the learning rate $\alpha_{prev} \in [0, 1e-6, 1e-5, 1e-4]$ for the previous classifier in adversarial tuning. More details of our experimental settings and comparison baselines are included in Appendix \ref{sec:exp-details} and \ref{sec:baselines}. 

\begin{figure}[t]
\centering
\includegraphics[width=0.9\columnwidth]{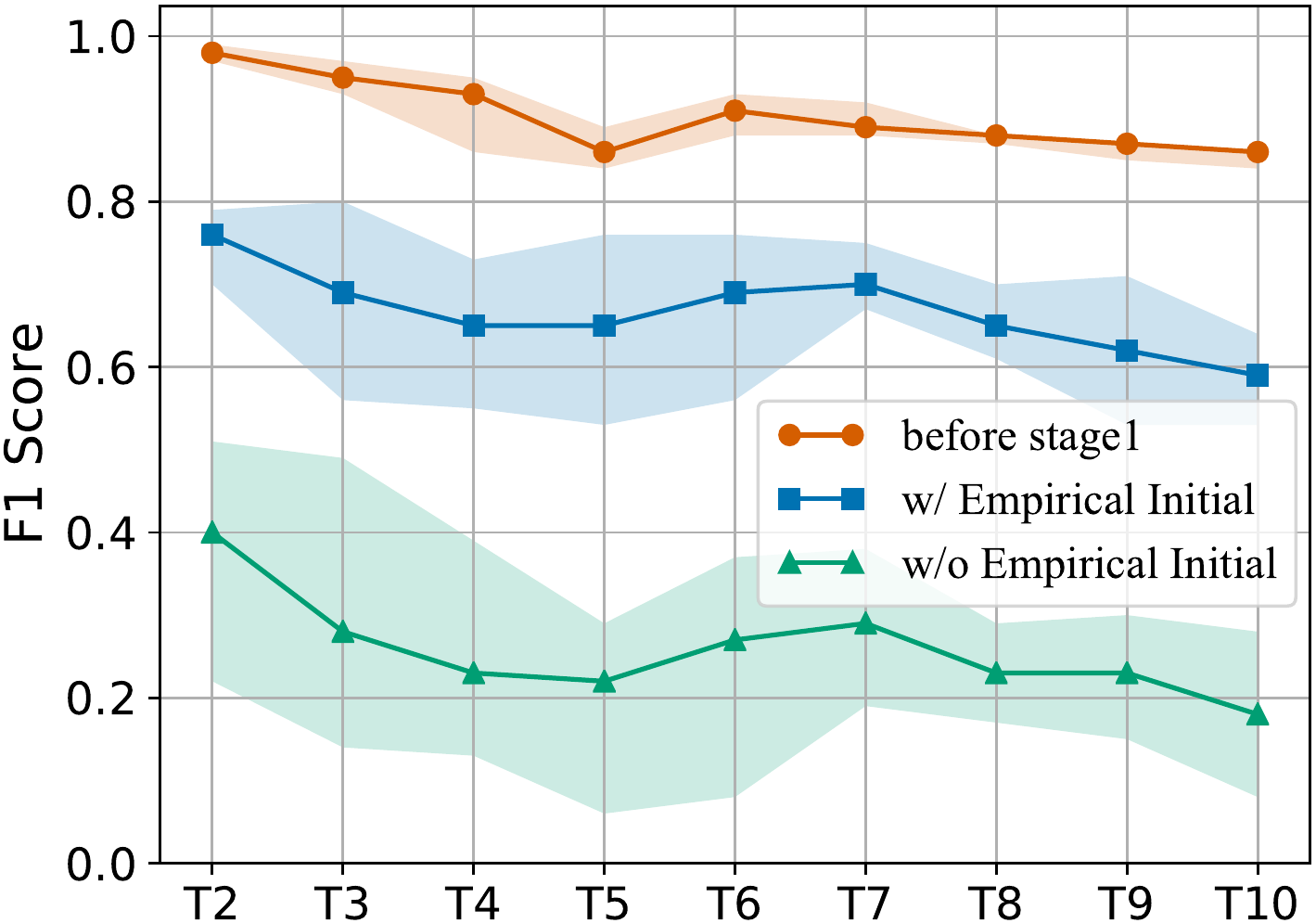}
\caption{F1 scores of previous relations in the training of stage 1. Results are obtained on the FewRel dataset.}
\label{fig:initial}
\end{figure}

\subsection{Main Results}
The performances of our proposed classifier decomposition framework and baselines on two datasets are shown in Table \ref{tab:main}. The results indicate that our framework is significantly superior to other baselines ($p<0.05$) and achieves state-of-the-art performance in the vast majority of settings. Besides, this framework is orthogonal to the data augmentation strategy proposed by \citet{Wang:2022b}, which can further boost the model's performance.

\subsection{Analysis}
\paragraph{Effectiveness of Empirical Initialization}
As shown in Figure \ref{fig:initial}, after stage 1, the model's performance on previous relations is consistently improved with our proposed empirical initialization strategy on all tasks, which indicates that the bias in previous relations is effectively alleviated.

\begin{figure}[t]
\centering
\includegraphics[width=0.9\columnwidth]{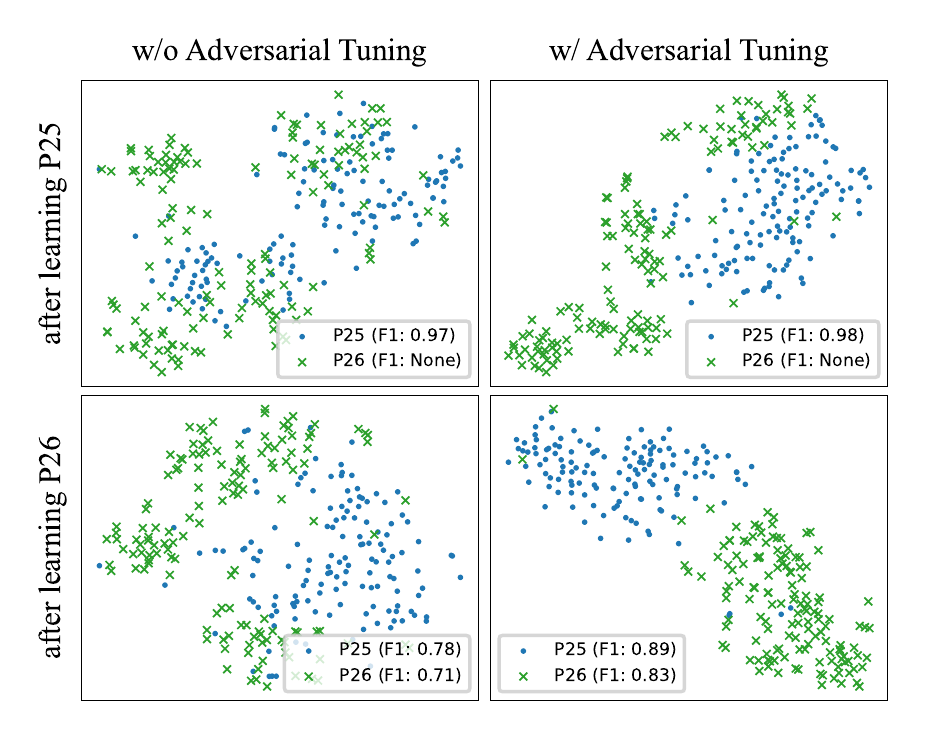}
\caption{The instance representations belonging to P25 (``mother'') and P26 (``spouse'') after learning P25 and P26, respectively.}
\label{fig:rep}
\end{figure}

\paragraph{Effectiveness of Adversarial Tuning} We utilize t-SNE to visualize the representations of a analogous relation pair: P25 (``mother'') and P26 (``spouse'')\footnote{We include more cases in Appendix \ref{sec:reps_at}.}. As shown in Figure \ref{fig:rep}, with adversarial tuning, the representations of instances belonging to P25 and P26 is much more separable compared with those of the vanilla training strategy, which indicates that adversarial tuning indeed alleviates the representation bias and helps the model learn more robust representations.

\begin{table}[t]
\centering
\small
\begin{tabular}{l|cc}
\toprule
\textbf{Models} & \textbf{FewRel} & \textbf{TACRED}  \\
\midrule
Ours & 84.6 & 78.6 \\
\midrule
w/o Empirical Initialization & 84.3 & 78.4 \\
w/o Adversarial Tuning & 84.0 & 77.9 \\
w/o both & 83.5 & 77.6 \\
\bottomrule
\end{tabular}
\caption{Accuracy (\%) of models with different strategies.}
\label{tab:ablation}
\end{table}

\paragraph{Ablation Study}
We further conduct an ablation study of our proposed two enhanced strategies in Table \ref{tab:ablation}. The experimental results show a performance degradation with the ablation of both strategies, demonstrating the effectiveness of our proposed classifier decomposition framework.

\section{Conclusion}
In this work, we found that the vanilla training strategy adopted by most previous CRE models in the first training stage leads to two typical biases: \textit{classifier bias} and \textit{representation bias}, which are important factors causing catastrophic forgetting. To this end, we propose a simple yet effective classifier decomposition framework with two enhanced strategies to help models alleviate those biases at the first training stage. Experimental results on two benchmarks show that our framework consistently outperforms previous state-of-the-art CRE models, which indicates that the value of this training stage to CRE models may be undervalued. Further analysis shows the effectiveness of our proposed classifier decomposition framework.

\section*{Limitations}\label{subsec:limitation}
As a preliminary study, our proposed classifier decomposition framework focuses on the first training stage of CRE models with a lack of explorations on stage 2. Besides, more experiments can be conducted by combining our framework with previous leading CRE models, which we leave for future research. In addition, our work only focuses on strategies with the FFN layer. As the BERT encoder is the main component of CRE models, we call for more attentions to the research of improving encoder representations in the first training stage.

\section*{Acknowledgements}\label{subsec:acknowledgements}
We thank all the anonymous reviewers for their thoughtful and constructive comments. This paper is supported by the National Key Research and Development Program of China 2020AAA0106700 and NSFC project U19A2065.

\bibliography{anthology,custom}
\bibliographystyle{acl_natbib}

\clearpage

\appendix
\section{Experiments Details}
\subsection{Datasets}
\label{sec:datasets}
Following previous work \cite{Han:2020, Cui:2020, Hu:2022, Wang:2022a}, our experiments are conducted upon the following two standard benchmarks with the training-test-validation split ratio set to 3:1:1.

\paragraph{FewRel} \cite{Han:2018} It is a RE benchmark dataset originally proposed for few-shot learning. The dataset contains 100 relations, each with 700 instances. Following the previous work \cite{Wang:2019, Han:2020, Zhao:2022}, we use the original training and validation set of FewRel, which contains 80 relations.

\paragraph{TACRED} \cite{Zhang:2017} It is a large-scale RE dataset containing 42 relations (including \textit{no\_relation}) and 106,264 samples, which is constructed on news networks and online documents. Following \citet{Cui:2020}, we removed \textit{no\_relation} in our experiments. The number of training samples for each relation is limited to 320 and the number of test samples of relation to 40.

\subsection{Experimental Details}
\label{sec:exp-details}
Following previous work \cite{Han:2020, Cui:2020}, we use bert-base-uncased as our encoder and Adam as our optimizer. We set the learning rate $1e-3$ for non-BERT modules and $1e-5$ for the BERT module, if not specified. The batch size of training is 32. The memory size of each task is 10. The training epoch for stage 1 and stage 2 are set to 10 for FewRel and 8 for TACRED. Our experiments are conducted on a single NVIDIA 3090 GPU.

\subsection{Baselines}
\label{sec:baselines}
We compare our proposed framework with the following baselines in our experiments:

\begin{itemize}
    \item \textbf{EA-EMR} \cite{Wang:2019} proposes a memory replay and embedding alignment mechanism to alleviate the problem of catastrophic forgetting.
    \item \textbf{EMAR} \cite{Han:2020} constructs a memory activation and reconsolidation mechanism to alleviate the catastrophic forgetting.
    \item \textbf{CML} \cite{Wu:2021} introduces curriculum learning and meta-learn to alleviate order sensitivity and catastrophic forgetting in CRE.
    \item \textbf{RPCRE} \cite{Cui:2020} proposes relation prototypes and a memory network to refine sample embeddings, which effectively retains the learned representations in CRE;
    \item \textbf{CRL} \cite{Zhao:2022} proposes to utilize contrastive learning and knowledge distillation to alleviate catastrophic forgetting.
    \item \textbf{CRECL} \cite{Hu:2022} introduces prototypical contrastive learning to ensure that data distributions of all CRE tasks are more distinguishable to alleviate catastrophic forgetting.
\end{itemize}

\section{Learning rate in Adversarial Tuning} 
We search various learning rates for previous classifier in \textit{adversarial tuning}. The results shown in Table \ref{tab:lr} indicate that the rate of $1e-5$ performs best, which is on par with that of the BERT encoder. It is worth noting that completely freezing the previous classifier (i.e., $lr=0$) leads to performance degradation since the model easily overfits on the fixed output distribution of the previous classifier.

\begin{table}[t]
\centering
\small
\begin{tabular}{l|ccccc}
\toprule
\textbf{Learning Rate} &$0$ & $1e-6$ & $1e-5$ & $1e-4$\\
\midrule
\textbf{FewRel} & 82.2 & 84.0 & \textbf{84.6} & 83.9 \\
\textbf{TACRED} & 76.5 & 77.8 & \textbf{78.6} & 78.0\\
\bottomrule
\end{tabular}
\caption{Accuracy (\%) of models with various learning rates.}
\label{tab:lr}
\end{table}

\section{Case Study}
\label{sec:cases}
\begin{figure}[t]
\centering
\includegraphics[width=1.0\columnwidth]{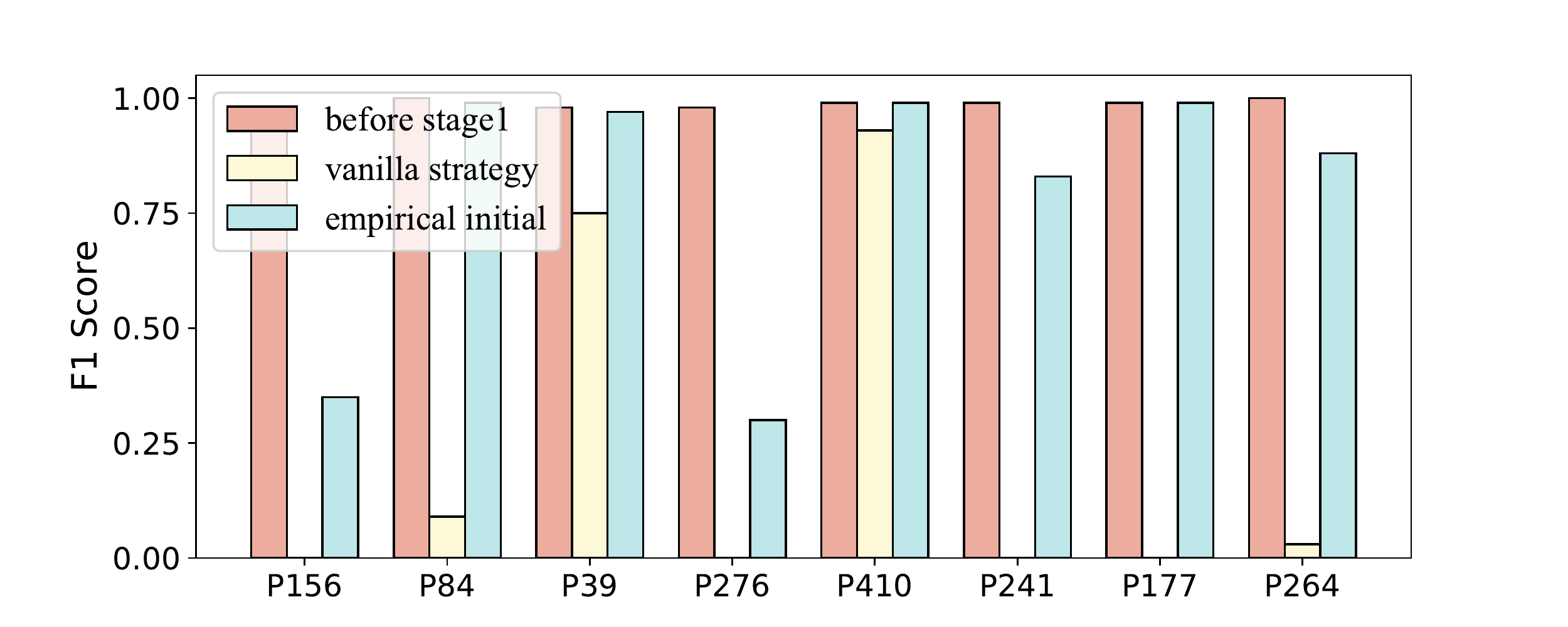}
\caption{A case of empirical initialization on the FewRel dataset.}
\label{fig:initial_case}
\end{figure}

We show a specific case including F1 scores of previous relations after the first training stage of the task $T_2$. As shown in Figure \ref{fig:initial_case}, significant improvement in scores of previous relations is shown with the empirical initialization strategy, indicating its effectiveness.

\section{Representations with Adversarial Tuning}
\label{sec:reps_at}
\begin{figure*}[t]
\centering
\includegraphics[width=1.0\textwidth]{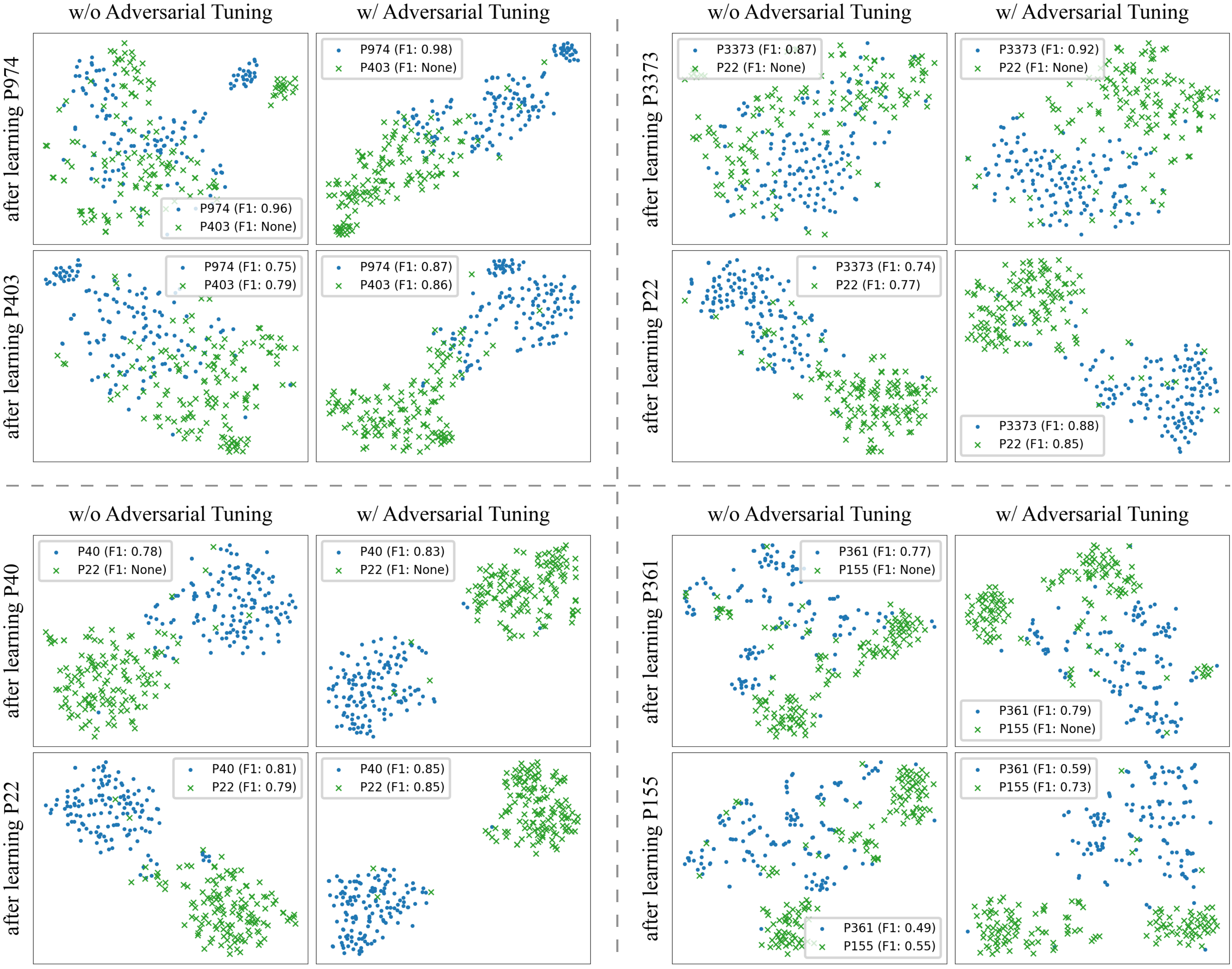}
\caption{More cases of robust representation learning with adversarial tuning.}
\label{fig:at_cases}
\end{figure*}

We illustrate more cases in Figure \ref{fig:at_cases} to show the effectiveness of adversarial tuning strategy.

\end{document}